\documentclass[letterpaper, 10 pt, journal, twoside]{IEEEtran}
\usepackage{float}
\usepackage{graphicx}
\usepackage{amssymb}

\usepackage{amsmath}
\usepackage{mathtools}  
\usepackage{diagbox}
\usepackage{multirow}
\usepackage{multicol}
\usepackage{booktabs}
\usepackage{hyperref}
\usepackage[font=small,labelfont=bf]{caption}
\usepackage{subcaption}
\usepackage{academicons}
\usepackage{xcolor}
\usepackage{tikz}
\usepackage{siunitx}

\renewcommand{\baselinestretch}{0.96}
\ifCLASSINFOpdf
\else
\fi
\newtheorem{problem}{Problem}

\newcommand{\customsubsection}[1]{\vspace{4px}\noindent \textbf{#1}}

\newcommand{\note}[1]{{\color{orange}* #1}}
\newcommand{\eatnotes}[0]{\renewcommand{\note}[1]{}}
\eatnotes

\newcommand{\robotposition}{\boldsymbol{r}}
\newcommand{\expnumber}[2]{{#1} \times 10^{#2}}
\newcommand{\dragforce}{\mathbf{f}^{drag}}
\newcommand{\wind}{\boldsymbol{w}}
\newcommand{\pressure}{p}
\newcommand{\trajtrackforce}{\boldsymbol{f}^{tt}}
\newcommand{\rlforce}{\boldsymbol{f}^{rl}}
\newcommand{\statespace}{{\mathcal{S}}}
\newcommand{\actionspace}{{\mathcal{A}}}
\newcommand{\replaybuffer}{{\mathcal{D}}}
\newcommand{\expected}{{\mathbb{E}}}
\newcommand{\entropy}{{H}}
\newcommand{\cat}[2]{#1 \; \| \; #2}
\newcommand{\neighborhood}[1]{\mathcal{N}_{#1}}
\newcommand{\workspace}{\mathcal{W}}
\newcommand{\graph}{\mathcal{G}}
\newcommand{\edges}{\mathcal{E}}
\newcommand{\nodes}{\mathcal{N}}
\newcommand{\features}{\boldsymbol{H}}
\newcommand{\graphdef}{\graph = \left( \nodes, \edges, \features \right)}
\newcommand{\adjacency}{\boldsymbol{A}_{adj}}
\newcommand{\actorparscaps}{\boldsymbol{\Theta}}
\newcommand{\actorpars}{\boldsymbol{\theta}}

\newcommand{\policy}{\boldsymbol{\pi}^{\actorpars}}

\DeclareRobustCommand{\orcidicon}{%
	\begin{tikzpicture}
	\draw[color={rgb:red,166;green,206;yellow,57}, fill={rgb:red,166;green,206;yellow,57}] (0,0) 
	circle [radius=0.16] 
	node[white] {{\fontfamily{qag}\selectfont \tiny ID}};
	\draw[white, fill=white] (-0.0625,0.095) 
	circle [radius=0.007];
	\end{tikzpicture}
	\hspace{-2mm}
}
\newcommand{\orcid}[1]{\href{https://orcid.org/#1}{\textcolor[HTML]{A6CE39}{\orcidicon}}}

\begin{document}
%
\title{Learning to Navigate in Turbulent Flows with Aerial Robot Swarms: A Cooperative Deep Reinforcement Learning Approach}
%
%
%


\author{Diego Patiño$^{1}$\orcid{0000-0003-4808-8411}, Siddharth Mayya$^{2}$\orcid{0000-0001-5168-413X}, Juan Calderon$^{3}$\orcid{0000-0002-4471-3980}, Kostas Daniilidis$^{1}$\orcid{0000-0003-0498-0758} and David Saldaña$^{4}$\orcid{0000-0003-2442-4257}%
\thanks{Manuscript received: December 23, 2022; Revised Feb 20, 2023; Accepted April 24, 2023.}
\thanks{This paper was recommended for publication by Editor Jens Kober upon evaluation of the Associate Editor and Reviewers' comments. 
This work was supported by ARO MURI W911NF-20-1-0080, ONR N00014-17-1-2093, ONR N00014-22-1-2677.} 
\thanks{$^{1}$ D. Patiño and Kostas Daniilidis are with the 
GRASP Lab., University of Pennsylvania, PA, USA:
        {\tt\small \{diegopc, kostas \}@cis.upenn.edu}
        }\thanks{
$^{2}$ S. Mayya is with Amazon Robotics, North Reading, MA, USA: 
        {\tt\small mayya.siddharth@gmail.com}. This work is not related to Amazon. 
    }\thanks{
        $^{3}$ J. Calderon is with Universidad Santo Tomas, Colombia, and Bethune Cookman University, FL, USA: {\tt\small calderonj@cookman.edu}}%
\thanks{$^{4}$ D. Saldaña is with the Autonomous and Intelligent Robotics Laboratory (AIRLab), Lehigh University, PA, USA: \texttt{saldana@lehigh.edu}}  
\thanks{Digital Object Identifier (DOI): see top of this page.}
}

%
%

\markboth{IEEE Robotics and Automation Letters. Preprint Version. Accepted April, 2023}
{Patiño \MakeLowercase{\textit{et al.}}: Learning to Navigate in Turbulent Flows with Aerial Robot Swarms} 

%



\maketitle
\begin{abstract}
Aerial operation in turbulent environments is a challenging problem due to the chaotic behavior of the flow. This problem is made even more complex when a team of aerial robots is trying to achieve coordinated motion in turbulent wind conditions.
In this paper, we present a novel multi-robot controller to navigate in turbulent flows, decoupling the trajectory-tracking control from the turbulence compensation via a nested control architecture. Unlike previous works, our method does not learn to compensate for the air-flow at a specific time and space. Instead, our method learns to compensate for the flow based on its effect on the team. This is made possible via a deep reinforcement learning approach, implemented via a Graph Convolutional Neural Network (GCNN)-based architecture, which enables robots to achieve better wind compensation by processing the spatial-temporal correlation of wind flows across the team. Our approach scales well to large robot teams ---as each robot only uses information from its nearest neighbors---, and generalizes well to robot teams larger than seen in training.
Simulated experiments demonstrate how information sharing improves turbulence compensation in a team of aerial robots and demonstrate the flexibility of our method over different team configurations.


\end{abstract}

\begin{IEEEkeywords}

Swarm Robotics, Reinforcement Learning, Wind Turbulence, Machine Learning for Robot Control, Graph Neural Networks.
\end{IEEEkeywords}


\section{Introduction}
\IEEEPARstart{A}{erial} vehicles naturally have to operate in environments with windy conditions. The wind field directly affects the vehicle's motion, potentially leading it outside its desired trajectory or even to crash. Navigating in windy conditions is even more difficult when air-flow is turbulent, presenting a chaotic behavior with hard-to-predict changes in pressure and flow velocity.   
This challenge is exacerbated in aerial multi-robot scenarios where a team of robots has to perform coordinated tasks which might require staying within communication range without colliding with one another. However, operating multi-robot systems in turbulent environments is highly relevant to reducing delivery and transportation delays, as well as supporting search and rescue operations during natural disasters from storms, tornadoes, and hurricanes. \par

The existing robotics literature has studied the problem of navigation flows, relying on assumptions to make the problem tractable. While some approaches assume a known (static or dynamic) wind field, e.g.,~\cite{garau2005path, bakolas2010time, otte2016}, other methods learn an association between a \emph{location} in the environment and the effect of the flow~\cite{bisheban2018geometric, montella2014reinforcement}. These are relevant limitations because it does not allow the robots to reuse their learned information in turbulent flows where such associations are constantly evolving or being faced with a new or unknown environment.


\begin{figure}
    \setlength\belowcaptionskip{-1.5\baselineskip}
     \centering
     \begin{subfigure}[b]{0.23\textwidth}
        \setlength\belowcaptionskip{0.0\baselineskip}
         \centering
         \includegraphics[width=\textwidth, trim=0.4cm 0.0cm 0.0cm 0.0cm,clip]{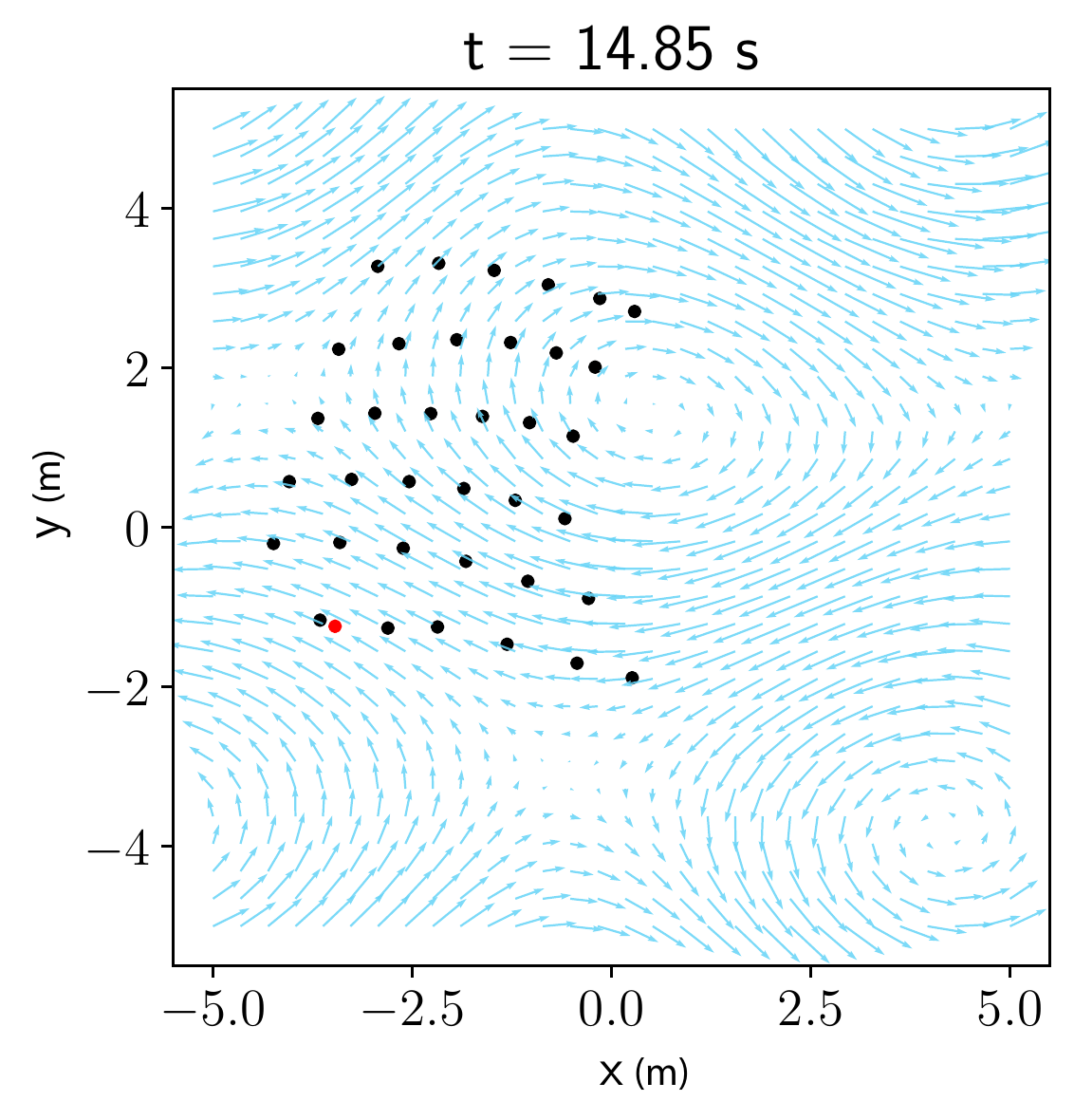}
     \end{subfigure}
     \hfill
     \begin{subfigure}[b]{0.23\textwidth}
        \setlength\belowcaptionskip{0.0\baselineskip}
         \centering
         \includegraphics[width=\textwidth, trim=0.4cm 0.0cm 0.0cm 0.0cm,clip]{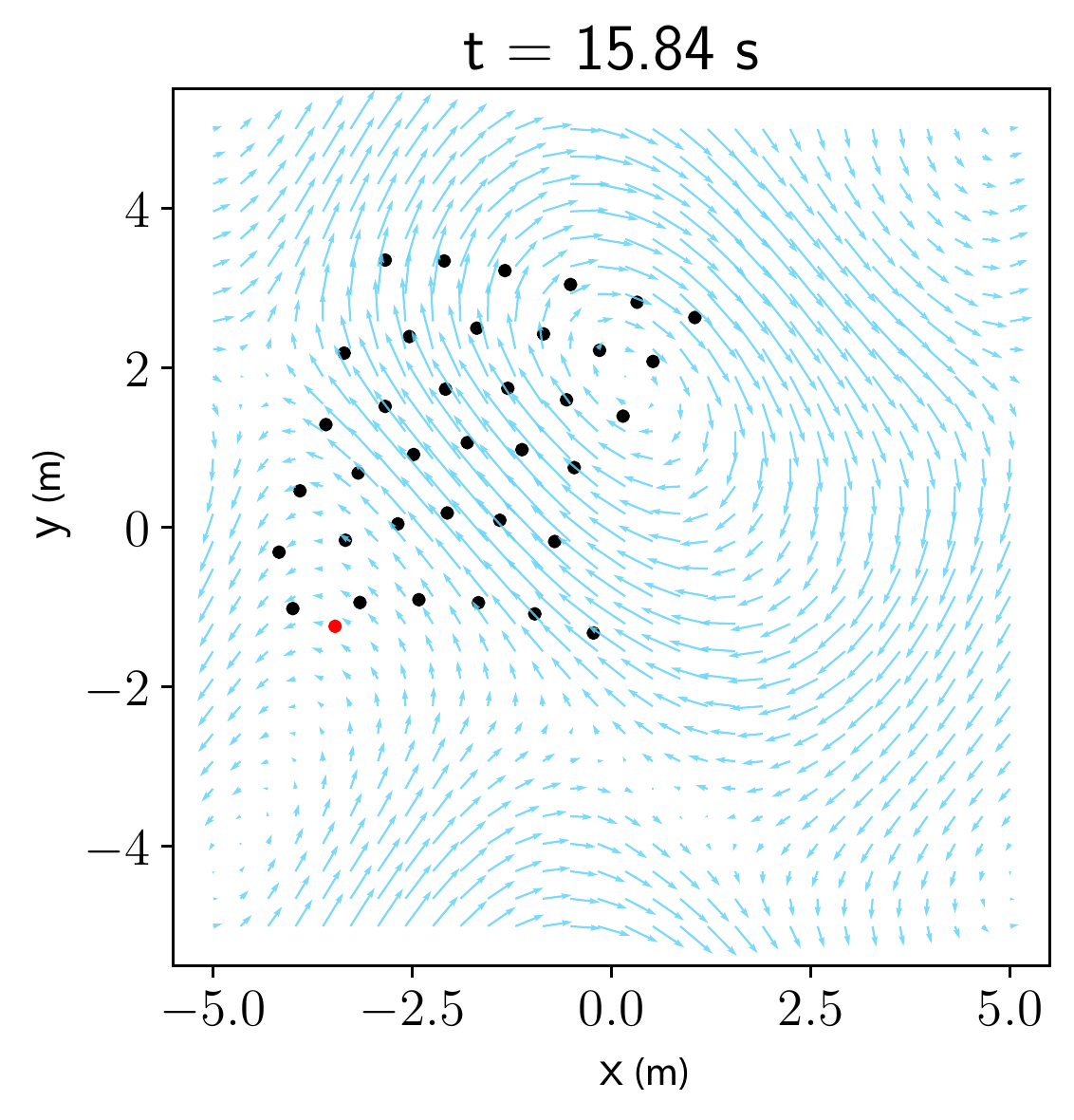}
     \end{subfigure}
     
     \begin{subfigure}[b]{0.23\textwidth}
       \setlength\belowcaptionskip{0.0\baselineskip}
        \centering
        \includegraphics[width=\textwidth, trim=0.4cm 0.0cm 0.0cm 0.0cm,clip]{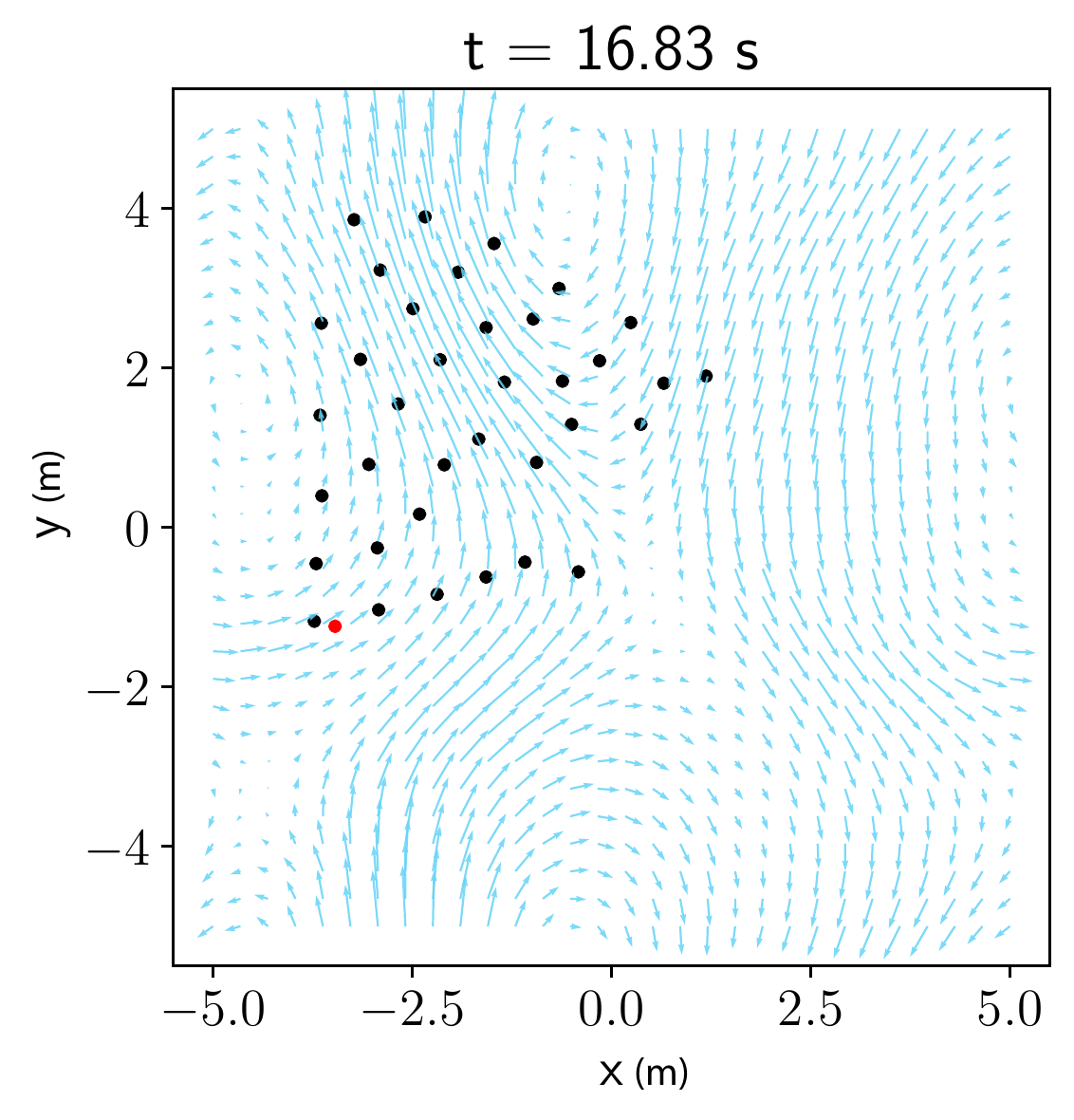}
     \end{subfigure}
     \hfill
     \begin{subfigure}[b]{0.23\textwidth}
       \setlength\belowcaptionskip{0.0\baselineskip}
        \centering
        \includegraphics[width=\textwidth, trim=0.4cm 0.0cm 0.0cm 0.0cm,clip]{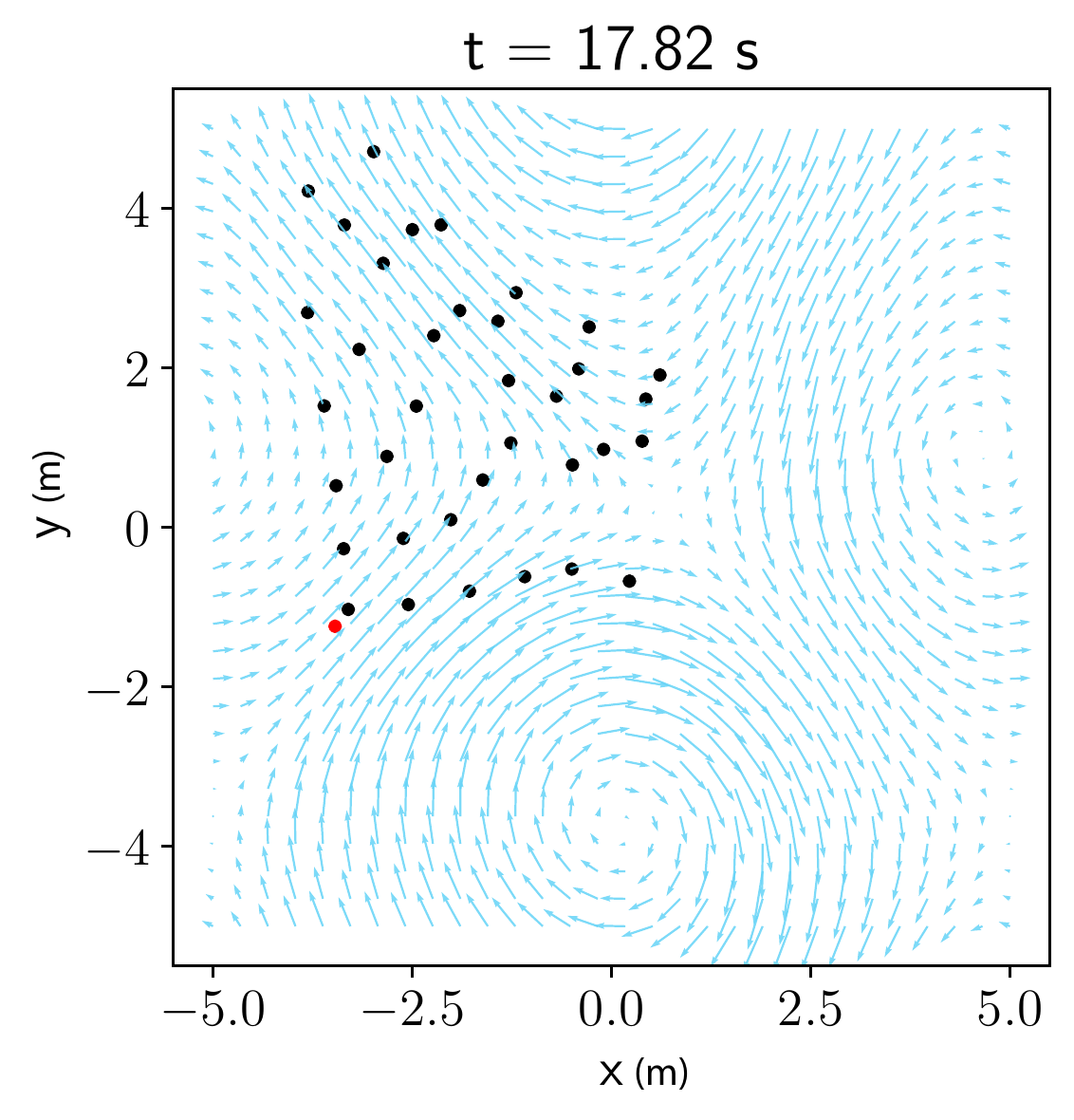}
     \end{subfigure}
        \caption{A team of 36 robots navigating in turbulent wind. The robots are trying to maintain a square formation using only a trajectory-tracking controller. Blue arrows show the wind vector field. The red dot shows the target location of the bottom-left robot in the formation. The X-axis and y-axis units are in meters.}
        \label{fig:scenario}
\end{figure}

In Fig.~\ref{fig:scenario}, we show an aerial multi-robot system operating in a turbulent wind flow. This figure illustrates a key observation: \emph{sensory information sharing} can provide valuable information to improve the robots' turbulence compensation in the absence of predictive wind-flow maps. For example, the approach of a new wind front could be detected by a robot, which can then relay pertinent information to other robots to better compensate the wind. This essentially occurs because the rapid fluctuations in wind velocity and direction inherent to turbulent winds are spatio-temporally correlated across the region.

The primary contribution of this paper is a novel method for trajectory tracking in turbulent flows using multiple aerial vehicles equipped with sensors to measure wind pressure and relative distance to other robots. Specifically, our method leverages structured information sharing over a graph where robots represent nodes and communication between robots represents edges. To ensure generality over qualitatively different turbulent flows, we develop a deep reinforcement learning approach, implemented via a Graph Convolutional Neural Network (GCNN). Our approach learns to fuse and transform sensory information received from neighbors~\cite{gama2020graphs} in order to compensate for wind forces. \par 

Crucially, our method does not need to learn to map between a specific location and the wind flow. Instead, it leverages spatio-temporal correlations (as described by the Navier-Stokes equations~\cite{foias_manley_rosa_temam_2001}) in wind flow between team members. Our method ensures that the learned information will not be associated with a specific training environment or trajectory. Furthermore, this ensures a decoupling between the nominal trajectory tracking controller and the controller for turbulence compensation. \par 
Our approach is scalable due to the use of the GCNN because each robot only uses the information from its on-board sensors and the information of its neighbors in the communication graph. Our experiments demonstrate this scalability as well as the efficacy of the proposed approach. These experiments also offer insights into how the learned models leverage shared information among the robots for effective turbulence compensation.\par




 

\customsubsection{Related Work:} \label{sec:related_work} 
The original robotic navigation problem in windy environments was proposed by Zermelo in 1931 \cite{zermelo1931navigationsproblem}.
When modeling the flow as a vector field, some works assume that the flow is known and quasi-static, i.e., does not change in time and space. These works focus on developing planning methods for static vector fields \cite{garau2005path, bakolas2010time},
and spatio-temporal dynamic fields \cite{otte2016}.
However, those methods rely on knowing the vector field at the planning stage, which is unpredictable for turbulent flows.\par 
For unknown static flows, the works in \cite{chang2017motion, allison2019estimating} design robot navigation strategies that drive the robot to sweep the environment and create a map of the flow. 
In \cite{escareno2013trajectory}, the authors designed an adaptive controller for a quadrotor that models the flow as two parts: 1) a time-varying vector field that can be estimated and 2) an unknown speed-bounded flow that is assumed as noise.
Flow prediction is also studied and implemented in realistic settings \cite{rodriguez2017small, donnell2018wind}. Similar to the aforementioned works, however, they involve a large number of samples of the environment. \par

For unknown dynamics of the flow, learning approaches have shown promising results. 
A safe learning approach for a quadrotor is presented in \cite{wang2018safe}. Assuming the flow is static, the robot starts in a safe region that can be expanded as the learning process evolves. The work in~\cite{bisheban2018geometric} presents an adaptive flight control that learns how to track a given trajectory on a static flow. A reinforcement learning approach to navigate a static wind field is presented in \cite{montella2014reinforcement}. \par 

As discussed in the introduction, our method does not need to create associations between locations in the environment and the wind flow. Towards this end, we leverage Graph Convolutional Neural Networks (GCNNs)~\cite{gama2020graphs, ruiz2021graph}. They are effective at modeling associations within a graph and have been applied in a wide range of fields, including multi-robot coordination and decision making e.g.,~\cite{blumenkamp2022framework, gosrich2022coverage}.

\section{Problem Statement}
\label{sec:problem}
\customsubsection{Robot Team: }
Consider a team of  $n$ aerial robots, denoted by the set $\mathcal{V}=\{1,...,n\}$. 
Assuming that all robots are at the same height, we analyze their location and motion on the plane.
The position of each robot $i\in\mathcal{V}$ is denoted by  $\robotposition_i\in\mathbb{R}^2$. 
We define the state vector by the position and the velocity of the robot, i.e, $\boldsymbol{x}_i=[\robotposition_i^\top, \dot{\robotposition}_i^\top]^\top$.
We assume all robots are homogeneous and have the same mass $m$.
%
\note{Sensors and actuators:}
Each robot $i$ can use its local sensors to estimate its state as well as select variables of the environment. 
\note{Additionally, the robot can measure the pressure at its location, denoted by $\rho_i:=\rho_i(t, \mathbf{p}_i)$.}
Each robot can generate a force vector $\boldsymbol{f}_i\in \mathbb{R}^2$ as control input, i.e.,
\begin{equation}
{\boldsymbol{f}}_i=\boldsymbol{u}_i.    
\label{eq:u}
\end{equation}
For this formulation, our aerial vehicles can be a fully actuated hexarotor~\cite{hexa2015franchi} or an under-actuated quadrotor that tilts to generate a force in any direction~\cite{Lee2010GeometricSE3}. 
Each robot $i$ can exchange messages with its $k$ nearest robots 
denoted by $\mathcal{N}_i$.
At every time step, each robot communicates its state and information from on-board sensors. 

\customsubsection{Wind field:}
Our robot team operates in a windy environment $\workspace\subset \mathbb{R}^2$. We represent the wind's velocity at a time $t$ and a location $\robotposition_i \in \workspace$ as a vector-valued function $\wind:\mathbb{R}_{\geq 0}\times\mathbb{R}^2 \rightarrow \mathbb{R}^2$. The vector field follows the dynamics of a fluid, described by the incompressible Navier-Stokes equations~\cite{foias_manley_rosa_temam_2001}
\begin{align}
    \nabla \cdot \wind &= 0 \nonumber \\
    \dot{\wind} + \wind \cdot \nabla \wind &= - \nabla \pressure + \frac{1}{Re}\nabla^2 \wind,
\label{eq:navier}
\end{align}
where $Re$ is the flow's Reynolds number, and $\pressure$ is the scalar pressure field. The Reynolds number measures the ratio between inertial and viscous forces. It characterizes flow patterns in a fluid, e.g., at low $Re$, flows tend to be laminar, while at high $Re$ flows tend to be turbulent. In this work, we focus on turbulent environments with high Reynolds numbers ~\cite{gilbert-kawai_wittenberg_2014}, $Re \geq \expnumber{4}{3}$, in the flow dynamics \eqref{eq:navier}. Note that this type of turbulent environment has not been explored in the mobile robotics literature.


As a robot moves through the air, the wind exerts a drag force on the robot in the fluid's direction~\cite{batchelor2000introduction}. We compute the drag force as 
\begin{equation}
\dragforce = \frac{1}{2} \rho \|\wind\|^2 C_d \: A \: \hat{\wind},
\label{eq:drag}
\end{equation}
where $\rho$ is the air density, the operator $\| \cdot \|$ is the 2-norm, $C_d$ is the robot's drag coefficient, $A$ is the cross-sectional area, and $\hat{\wind}$ is a unit vector in the direction of ${\wind}$. In this context, the reference area is the orthogonally projected frontal area, i.e., the object's visible area as seen from a point on its line of travel. We assume that the drag coefficient and the air density are constant.

\customsubsection{Sensors:}
The robots in our team do not know the wind field nor any of the coefficients in \eqref{eq:drag}. However, they can use their equipped sensors and noisy measurements to gather information about their surroundings.
Each robot is equipped with a pressure sensor, a location sensor, and an inertial measurement unit (IMU). The IMU estimates the robot's linear velocity. The robot can measure the relative distance to their $k$-nearest neighbors using any relative location system, e.g., camera, LIDAR, sonar, or time of flight (ToF) sensor.

\note{Newton Dynamics:}
\customsubsection{Robot dynamics:}
The robot's actuation and the turbulent wind generate linear forces that determine the robot's motion. We model the dynamics of the $i$th robot using Newton's equation, 
\begin{equation}
    m\ddot{\robotposition}_i = {\boldsymbol{u}}_i + \dragforce_i.
    \label{eq:newton}
\end{equation}
\customsubsection{Trajectory tracking:}
The goal for the robot~$i$ is to follow a given trajectory $\boldsymbol{x}_i^d(t)=[{\robotposition_i^{d\top}}(t), {\dot{\robotposition}_i^{d\top}}(t)]^\top$, specified by a desired location $\robotposition^d_i$ and desired velocity $\dot{\robotposition}^d_i$ in a time interval $[0,T_f]$~\cite{Mellinger2011MinimumQuadrotors}. 
Assuming an environment without wind, i.e., $\dragforce_i=\boldsymbol{0}$ in~\eqref{eq:newton}, we can use a classical trajectory-tracking approach that provides exponential stability \cite{khalil2002nonlinear} based on a feed-forward controller,
\begin{equation}
    \boldsymbol{u}^{tt}_i = \boldsymbol{K}_p ({\robotposition}^d_i - \robotposition_i) + \boldsymbol{K}_d (\dot{\robotposition}^d_i - \dot{\robotposition}_i)
    + \ddot{\robotposition}^d_i
    ,
    \label{eq:u_tt}
\end{equation}
where $\mathbf{K}_p$ and $\mathbf{K}_d$ are the diagonal gain matrices. The main challenge here is that $\dragforce_i$ is not negligible and can drive the robot far away from the given trajectory, thereby making the dynamical system in \eqref{eq:newton} unstable.

\note{Problem:}
\customsubsection{Objective:}
Our objective is to allow the robot team to track a trajectory while operating in a dynamic, turbulent wind field. 
We, therefore, need to solve the following problem:
\begin{problem}
Given a set of $n$ robots and a trajectory that can be solved with a control policy $\boldsymbol{u}^{tt}_i$, which does not consider turbulence, find a control input $\boldsymbol{u}_i$ such that the robots can perform the given task in a turbulent environment.
\end{problem}
Our key insight is that although the robots do not know the wind field, each can share its state and sensor measurements with neighboring robots. Sharing information allows each robot to increase its knowledge about the working environment, leading to an action policy that effectively compensates for the wind's drag force. 

Note that our approach is independent of the trajectory tracking because we aim to learn the wind patterns independently of the trajectory-tracking controller. 

\section{Deep Reinforcement Learning Method} \label{sec:rl}
\begin{figure}[t]
    \setlength\belowcaptionskip{-1.5\baselineskip}
    \centering
    \includegraphics[width=\linewidth]{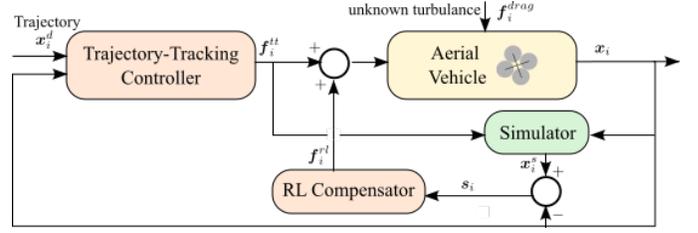}
    \caption{Control diagram of our proposed method.}
    \label{fig:control}
\end{figure}

\customsubsection{Control Strategy:}
The key to our control strategy is decoupling the trajectory-tracking controller and the wind compensation. Trajectory-tracking controllers already show exponential convergence \cite{Lee2010GeometricSE3,khalil2002nonlinear}. However, convergence is not guaranteed when an external force from the wind is added to the dynamics as modeled in \eqref{eq:newton}. To overcome this limitation, we leverage Reinforcement Learning (RL) to design a second controller that compensates for wind disturbances. This new controller forms an inner control loop, as seen in Fig.~\ref{fig:control}, and assists the trajectory-tracking controller by helping it converge as if operating in a disturbance-free setting.

The  force generated by a robot is the combination of an RL-based 
wind compensation force $\rlforce_i $ and trajectory tracking force $\trajtrackforce_i$. So the total force generated by the robot is 
$\boldsymbol{u}_i = \rlforce_i + \trajtrackforce_i$. Substituting the total force in \eqref{eq:newton}, we obtain
\begin{equation}
    m\ddot{\robotposition}_i = \rlforce_i + \trajtrackforce_i + \dragforce_i.
    \label{eq:forces}
\end{equation}
We set the trajectory-tracking force to be the control's action from \eqref{eq:u_tt}, such that $\trajtrackforce = \boldsymbol{u}^{tt}_i$. 

The purpose of the $\rlforce_i$ is to compensate for the
effect of the wind flow, thereby allowing the robots to track their desired trajectory. To this end, let $\actionspace$ be the action space and $\statespace$ the state space in the RL context. We use a Deep-RL policy -- $\policy_i(\boldsymbol{a}_i | \boldsymbol{s}_i)$ -- to compute a wind compensation action for each robot. We model the policy with a deep neural network with parameters $\actorpars$, conditioned on a set of observed variables $\boldsymbol{s}_i \in \statespace$. Then, we set $\rlforce = \boldsymbol{a}_i$ where $\boldsymbol{a}_i \sim \policy_i(\boldsymbol{a}_i | \boldsymbol{s}_i)$.

We set the action space $\actionspace$ to be $[-f^{max}_{rl}, f^{max}_{rl}]^2 \subset \mathbb{R}^{2}$, representing a two-dimensional bounded force. Unlike classical RL methods, the $i$th robot's policy depends on all the states in the robotic team rather than just $\boldsymbol{s}_i$. This allows our method to use information across robots. Later in this section, we will offer a precise definition of $\statespace$ and the information-sharing architecture of our RL method.

\customsubsection{Soft Actor-Critic:} We learn $\policy(\boldsymbol{a}_{i} | \boldsymbol{s}_{i})$ using the Soft Actor-Critic algorithm (SAC). SAC is an off-policy Deep Reinforcement Learning (DRL) algorithm based on entropy regularization to trade off exploitation and exploration policies. SAC has demonstrated stability, sample-efficient learning, and optimal policy convergence \cite{Haarnoja2018SoftAO}. The SAC method optimizes $\policy_{i}$ by jointly maximizing its expected reward and its entropy~\cite{Haarnoja2018SoftAO,Haarnoja2019LearningTW}. Incorporating the entropy term into the RL framework casts an optimization problem of the form
\begin{equation}
{\pi}_i^* = \arg \max_{\pi} \underset{\tau \sim \pi}{\expected} \left[ { \sum_{t=0}^{\infty} \gamma_t \bigg( r(\boldsymbol{s}_i, \boldsymbol{a}_i, \boldsymbol{s}^{\prime}_i) + \alpha \entropy \left(\pi(\cdot|\boldsymbol{s}_i)\right) \bigg)} \right] ,\label{eq:entropy_reg_RL}
\end{equation}
where $\boldsymbol{s}^{\prime}_i$ is the state in the next time step after applying the action $\boldsymbol{a}_i$, $\alpha$ is the trade-off coefficient, $r$ is the reward signal, $\gamma$ is the discount factor, and $\entropy$ is the policy's entropy. The $\alpha$ values control the trade-off between the expected reward and entropy of the policy, balancing exploration and exploitation. Appropriate values of $\alpha$ accelerate the learning process towards the optimal policy and prevent convergence to local minima \cite{Haarnoja2018SoftAO}.

Following  \eqref{eq:entropy_reg_RL}, SAC uses a Deep Q-Learning strategy that incorporates $\entropy$ into a slightly modified version of the Bellman equation for the value function
\begin{equation}
V(\boldsymbol{s}_i) = \underset{\boldsymbol{a}_i \sim \pi}{\expected} \left[ {Q(\boldsymbol{s}_i, \boldsymbol{a}_i)} \right] + \alpha \entropy \left(\pi(\cdot| \boldsymbol{s}_i)\right)
\end{equation}
and the Bellman equation for the Q-function
\begin{equation}
Q(\boldsymbol{s}_i, \boldsymbol{a}_i) = \underset{\boldsymbol{s}^{\prime}_{i} \sim P}{\expected} \left[ {r(\boldsymbol{s}_i, \boldsymbol{a}_i ,\boldsymbol{s}^{\prime}_{i}) + \gamma V(\boldsymbol{s}^{\prime}_{i})} \right],
\end{equation}
where $P$ is the probability distribution of the future state $s^{\prime}_i$.

In practice, SAC estimates three functions: The policy (Actor) and two Q-functions (Critics). First, it approximates the policy as a Gaussian distribution $\policy \sim \mathcal{N}(\mu_{\actorpars}, \Sigma_{\actorpars})$. Both $\mu_{\actorpars}$ and $\Sigma_{\actorpars}$ are the outputs or a neural network parametrized with $\actorpars$ and optimized through gradient descent using the re-parametrization trick~\cite{Kingma2014AutoEncodingVB}. Similarly, SAC estimates two Q-functions $Q_{\actorpars_1}$ and $Q_{\actorpars_2}$ as neural networks with parameters $\actorpars_1$ and $\actorpars_2$, respectively. The Q-function networks train by minimizing the objective $J_{Q}(\actorpars_i)$
\begin{equation}
    \underset{(\boldsymbol{s}_i, \boldsymbol{a}_i, \boldsymbol{s}^{\prime}_i) \sim \replaybuffer}{\expected} \left[ \left( Q_{\actorpars_j}(\boldsymbol{s}_i, \boldsymbol{a}_i) - \left( r(\boldsymbol{s}_i, \boldsymbol{a}_i) + \gamma V_{\actorpars_1,\actorpars_2}(\boldsymbol{s}^{\prime}_i) \right) \right)^{2} \right]\label{eq:J_Q}
\end{equation}
over samples taken from a replay buffer $\replaybuffer = \statespace \times \actionspace \times \statespace$ of experience gathered during multiple episodes in the training process. The value function $V_{\actorpars_1,\actorpars_2}$ is implicitly defined through the Q-function and the policy, as stated in \cite{Haarnoja2019LearningTW}. Similarly, the objective for the Gaussian policy is given by
\begin{equation}
    J_{\pi}(\actorpars) = \underset{\boldsymbol{s}_i \sim \replaybuffer, \boldsymbol{a}_i \sim \policy}{\expected} \left[ \alpha \log \policy (\boldsymbol{a}_i | \boldsymbol{s}_i) - \min_{j \in \{1,2\}} Q_{\actorpars_j}(\boldsymbol{s}_i, \boldsymbol{a}_i) \right]\label{eq:J_pi}.
\end{equation}
Note that minimizing \eqref{eq:J_Q} is equivalent to finding the Q-function that best approximates the value function $V$. Analogous, minimizing \eqref{eq:J_pi} is equivalent to jointly maximizing the expected reward and the policy's entropy.

In this work, we adapt the SAC method to optimize the $i$th robot's policy conditioned on all the robot states in the team as opposed to a single agent state. 

\customsubsection{State space:}\label{subsec:state_space} Our approach does not focus on tracking the trajectory but on learning how to directly compensate for the disturbance experienced by the robots, such that the trajectory-tracking controller can operate freely. For this purpose, we integrate the dynamics in \eqref{eq:forces} to simulate the robot's dynamics under perfect conditions. In these conditions, there is no drag force and hence no need for wind compensation. Therefore, our RL approach's state $\boldsymbol{s}_i \in \statespace$ relates to how much the trajectory-tracking state $\boldsymbol{x}_i$ differs from a simulated state~$\boldsymbol{x}_i^{sim}$. Note that $\boldsymbol{x}_i^{sim}$ does not consider the wind effect.

In our method, we perform sampling and actuating periodically. Consequently, we assume that time is discrete, i.e., we use the variable, $\tau=0,1,2,...$ to represent discrete time steps. We use a constant step size $\Delta \tau$ small enough to apply our method in the dynamics equations in \eqref{eq:forces}.

\note{Simulation and state:}
Let us denote the trajectory-tracking state of the $i$th robot at a time step $\tau$ by $\boldsymbol{x}_i[\tau]$, and its simulated state by $\boldsymbol{x}_i^{sim}[\tau]$. 
Using Euler integration, we can predict the disturbance-free state~$\boldsymbol{x}_i^{sim}[\tau]$ using the past state $\boldsymbol{x}_{i}[\tau-1]$, and a trajectory-tracking action $\boldsymbol{u}^{tt}_{i}[\tau-1]$. We can write the discrete-time dynamics from \eqref{eq:forces} in matrix form, assuming $\dragforce = \rlforce = 0$, to compute the simulated state at $\tau$,
\begin{equation}
 \boldsymbol{x}_i^{sim}[\tau] = \boldsymbol{A} \boldsymbol{x}_{i}[\tau-1] + \boldsymbol{B} \boldsymbol{u}^{tt}_{i}[\tau-1],
 \label{eq:sim}
\end{equation}
where $$\boldsymbol{A} =\begin{bmatrix}
	\boldsymbol{1} & \Delta \tau~\boldsymbol{1} \\
	\boldsymbol{0} & ~~~\boldsymbol{1}
\end{bmatrix}, \text{ }
\boldsymbol{B} =\begin{bmatrix}
	\boldsymbol{0} & ~~~\boldsymbol{0} \\
	\boldsymbol{0}& \frac{\Delta \tau}{m} \boldsymbol{1}
\end{bmatrix}.$$ Then, the wind disturbance displacement vector is the difference between the current state $\boldsymbol{x}_i[\tau]$ and the simulated state~$\boldsymbol{x}_i^{sim}[\tau]$,
\begin{equation}
\boldsymbol{e}_i[\tau] = \boldsymbol{x}_i^{sim}[\tau] - \boldsymbol{x}_i[\tau].\label{eq:displacement_vector}
\end{equation}
As described in Sec. \ref{sec:problem}, the wind applies a drag force~$\dragforce_i$ on the robots. This force results from the pressure field gradient plus the friction forces due to air particles as described by \eqref{eq:navier}. Each robot takes noisy measurements of the pressure field $\pressure_i$ at its location to account for the effect of these forces.

Finally, we define the state vector~$\boldsymbol{s}_i$ for our RL method at each robot $i$, by concatenating the displacement vector $\boldsymbol{e}_i$, the pressure field value $\pressure_i$, and the robot's velocity $\dot{\robotposition_i}$ such that
\begin{equation}
    \boldsymbol{s}_i = \cat{\boldsymbol{e}_i}{\cat{\dot{\robotposition}_i}{\pressure_i}},\label{eq:state_vector_def}
\end{equation}
where $\cat{\cdot}{\cdot}$ is the concatenation operator. We include the robot's velocity because the drag force directly affects this quantity. During training, we add Gaussian noise to $\boldsymbol{s}_i$ to simulate real-world sensory noise, as discussed in Sec.~\ref{subsec:experimental_setup}.


\customsubsection{Graph Convolutional Neural Network Architecture:}
The wind flow dynamics in \eqref{eq:navier} reveal a spatio-temporal correlation for $\wind$, i.e., the wind velocity at a given location correlates with the wind velocities at nearby areas. Our proposed method takes advantage of the spatial correlation by enabling information sharing between the robotic team members. 

When we use multiple robots spatially distributed in $\workspace$, we form a sensing network that indirectly samples information about the effects of the wind on the robots. Consequently, we use this sensing network to improve the action that compensates for the drag force exerted on a robot~$i$ with the help of its neighbors~$\neighborhood{i}$. 

\begin{figure}[t]
    \setlength\belowcaptionskip{-1.5\baselineskip}
    \centering
    \includegraphics[width=\linewidth]{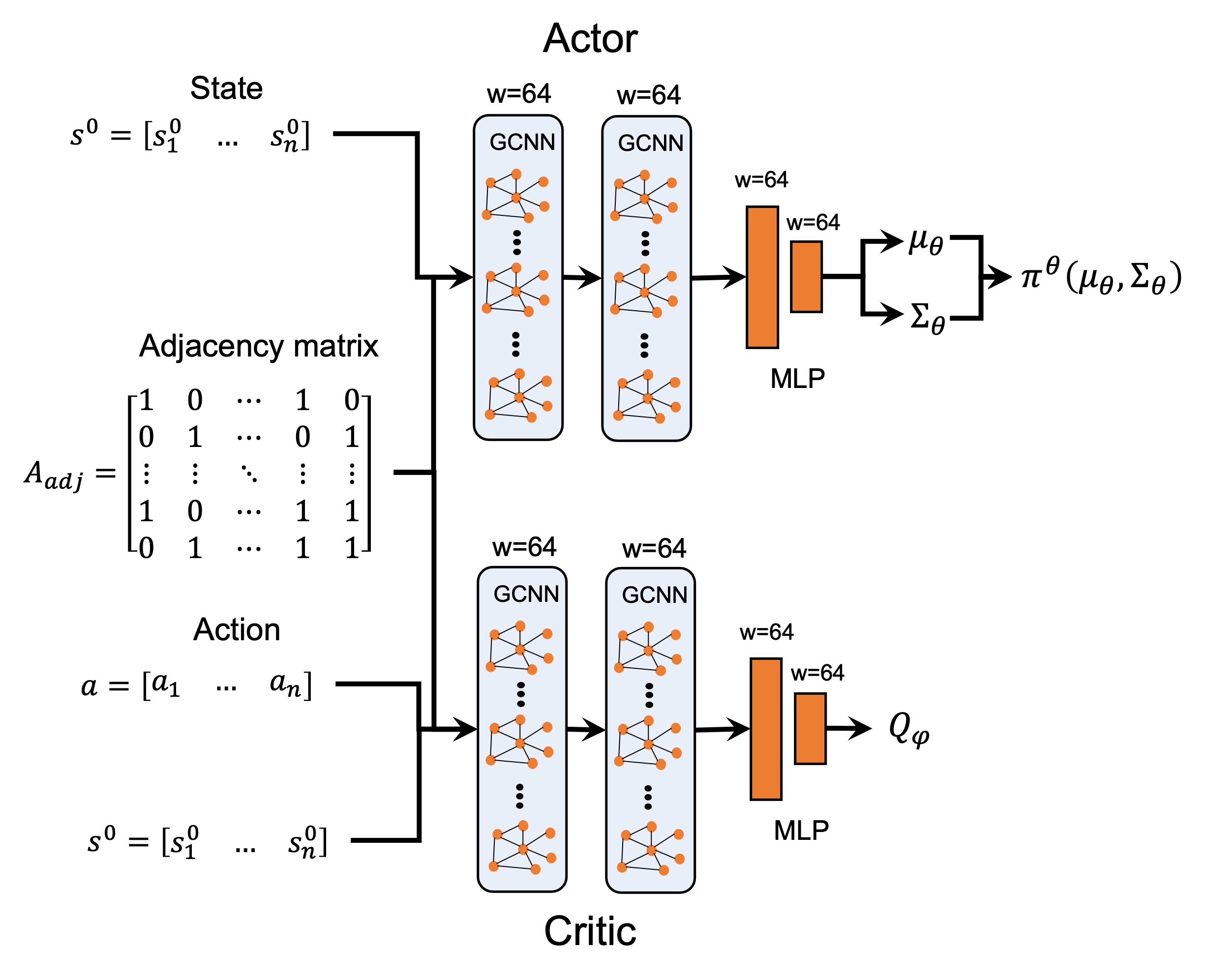}
    \caption{Team-level architecture of the actor and critic networks used within the proposed RL architecture.}
    \label{fig:arch}
\end{figure}

Since SAC was designed for a single agent, its actor's architecture is a multi-layer perceptron (MLP). An MLP acts only on the individual robot's states $\boldsymbol{s}_i$ to compute the robot's action $\boldsymbol{a}_i$. Hence, the MLP's architecture does not use information from other robotic team members.
To model this information exchange explicitly, we design the actor -- and the two critics -- as Graph Convolutional neural networks (GCNN)~\cite{Morris2019WeisfeilerAL}. A $L$-layered GCNN is a type of neural network that can process data represented as a graph $\graphdef$ with nodes $\nodes$, edges $\edges$, and a feature set $\features = \{\features^0, \ldots, \features^L\}$. In the context of this paper, the nodes represent robots, and the edges represent the information exchange between them. We present an overview of the full architecture for our GCNN-based actor and the critic in Fig.~\ref{fig:arch}. \par 

At a given layer $l \in [0,..., L]$, the network computes a feature vector for each robot $i$, denoted by $\boldsymbol{h}_{i}^{l}$, and organizes them into a $n \times c_l$ matrix
\begin{equation}
\boldsymbol{H}^{l} = [\boldsymbol{h}_{1}^{l}, .., \boldsymbol{h}_{n}^{l}]^\top.
\end{equation}
We compute $\boldsymbol{H}^{l}$ from the previous layer's features following
\begin{equation}
    \boldsymbol{H}^{l+1} = \sigma \left( \boldsymbol{H}^{l} \actorparscaps_1^{l} + \adjacency \boldsymbol{H}^{l} \actorparscaps_2^{l} \right),\label{eq:graph_conv_mat}
\end{equation}
where $\adjacency$ is the graph's adjacency matrix, $\actorparscaps_{1}^{l}$ and $\actorparscaps_{2}^{l}$ are learnable weight matrices of size $c_{l} \times c_{l + 1}$, and $\sigma(\cdot)$ is an element-wise non-linear activation function. 
We set the input features of the network to be a matrix containing all the robot's states defined in \eqref{eq:state_vector_def}, such that 
\begin{equation}
\boldsymbol{H}^{0} = [\boldsymbol{s}_1, ..., \boldsymbol{s}_n]^\top.
\end{equation}
The operation in \eqref{eq:graph_conv_mat} is a graph convolution operation where a robot's features are updated using information from its neighbors in the graph. However, this operation does not include information about the relative position $\robotposition_{ij}=\robotposition_{j} - \robotposition_{j}$ between robot~$i$ and its neighbor~$j$. Without the relative position, the robots do not know where the neighboring robots are located. This makes it difficult to approximate vector quantities such as the pressure gradient in \eqref{eq:navier}. To overcome this limitation, we incorporate the relative position into the convolution operator by concatenating $\robotposition_{ij}$ to the features at each layer right before the weighting and the neighbor aggregation. For simplicity, we will use the per-node notation of \eqref{eq:graph_conv_mat} to denote the convolution at each robot $i$. We define the layer's features at each robot as
\begin{equation}
    \boldsymbol{h}^{l+1}_{i} = \sigma \left( \actorparscaps^{l}_{1} \boldsymbol{h}^{l}_{i} +
\actorparscaps^{l}_{2} \sum_{j \in \neighborhood{i} } \left( \boldsymbol{h}^{l}_{j}\; ||\; \boldsymbol{r_{i,j}} \right) \right),\label{eq:graph_conv}
\end{equation}
where $\actorparscaps^l_{2}$ is now a $c_{l+1} \times (c_{l} + 2)$ matrix. The actor's GCNN architecture takes $\boldsymbol{H}^{0}$ and $\adjacency$ as inputs, and computes a latent vector representation $\boldsymbol{h}^{L}_{i}$ at the last layer $L$. To decode $\boldsymbol{h}^{L}_{i}$ into the robot's action, we pass $\boldsymbol{h}^{L}_{i}$ through an small MLP network. We split the MLP's output into $\boldsymbol{\mu}_i^{\actorpars}$ and $\boldsymbol{\Sigma}_i^{\actorpars}$, and we use them to parameterize $\policy$ as a normal distribution. Following \cite{Haarnoja2018SoftAO}, we set $\boldsymbol{\Sigma}_i^{\actorpars}$ to be a diagonal matrix. Finally, we use the policy to obtain the action~$\boldsymbol{a_i}$. 

Each of the critic's architecture follows a similar design with two small modifications since the critic is a function $Q: \statespace \times \actionspace \mapsto \mathbb{R}$. First, the critic's output is a single-value function instead of a probability distribution. To model its output properly, we modify the critic's MLP decoder to have a single output neuron rather than $\boldsymbol{\mu}_i^{\actorpars}$ and $\boldsymbol{\Sigma}_i^{\actorpars}$. Second, the input space of the critic architecture consists of the robot's action in addition to just the state. Consequently, the input to the critic's GCNN is a feature vector 
\begin{equation}
\boldsymbol{H}^{0^{\prime}} = \left[ \left(\cat{\boldsymbol{s}_1}{\boldsymbol{a}_1}), ..., (\cat{\boldsymbol{s}_n}{\boldsymbol{a}_n} \right) \right]^\top.
\end{equation}
In each architecture, we use a two-layer GCNN with $\mathrm{ReLU}$ as the non-linear activation function and two hidden layers of $64$ neurons per layer. The MLP decoders are two-layer networks of size $64$ and $16$, respectively. We add an extra output layer to the decoders to re-shape the network's output to the appropriate size for the actor or critics. The MLP's layers use $\mathrm{ReLU}$ as their activation function in the inner layers and a linear activation function for the output layer. 
Finally, the actor's output is squeezed into the range $[-1, 1]$ using a $\tanh$ function as described in the SAC formulation. In practice, we scale $\boldsymbol{a}_i$ by a preset factor of $\sqrt{2}f^{max}_{rl}$ representing the maximum force that the robots can generate, as discussed after~\eqref{eq:forces}.

Note that the depth of the GCNN is directly related to the robot network's bandwidth load. At a layer $l$, a robot has to communicate with its neighbors to compute $\boldsymbol{h}^{l}_{i}$. Since our architecture has only two layers, robot communication must only reach up to their 2-ring neighborhood. This property allows our model to scale to large formations since robot communication farther than the 2-ring is not required.

\customsubsection{Reward Signal:}
The final component of our proposed method is the reward signal. The reward signal at each step tells the SAC how well the robots compensate for the wind's drag force at a given step in the training process.

Recall that we expect the robot team to learn to operate the same as when there is no turbulence. Because turbulence affects the acceleration of the robots, the divergence between the expected simulated velocity at a robot $i$ and its actual velocity is an appropriate quantity to incorporate into our reward signal. We can do a similar analysis on the divergence between the simulated position and the actual position measured with the robot's instruments. These divergence quantities are captured into the displacement vector $\boldsymbol{e}_i$ in \eqref{eq:displacement_vector}. Hence, we define our reward function for our RL method as the L1-norm to weighted displacement,

\begin{equation}
{r}[\tau] = - \|  \boldsymbol{\beta} \odot  \boldsymbol{e}_i \|\label{eq:reward_signal},
\end{equation}
with $\odot$ the Hadamard product, and $\boldsymbol{\beta}$ a weight vector rating the importance of each component of $\boldsymbol{e}_i$ in the reward signal.

As the displacement vector between the simulated state and the actual state vector approaches the zero vector, the reward signal becomes less negative. Therefore, learning a policy that maximizes \eqref{eq:reward_signal} is equivalent to learning an action policy that compensates for the effect of the wind on the robots.

\section{Experiments}

We design three experiments to evaluate our method's performance. First, we show that our method allows robots to navigate turbulent wind regimes by independently compensating the wind and tracking the target's trajectory separately. Second, we show that our method is robust to changes in the robot team's configuration, such as neighborhood size and formation size. Third, we demonstrate that the advantages of our method arise from our GCNN-based RL strategy by ablating the GCNN and replacing it with an MLP.

\begin{figure}
    \setlength\belowcaptionskip{-1.5\baselineskip}
    \centering
    \includegraphics[width=\linewidth, trim=2.0cm 0.0cm 2.0cm 0.0cm,clip]{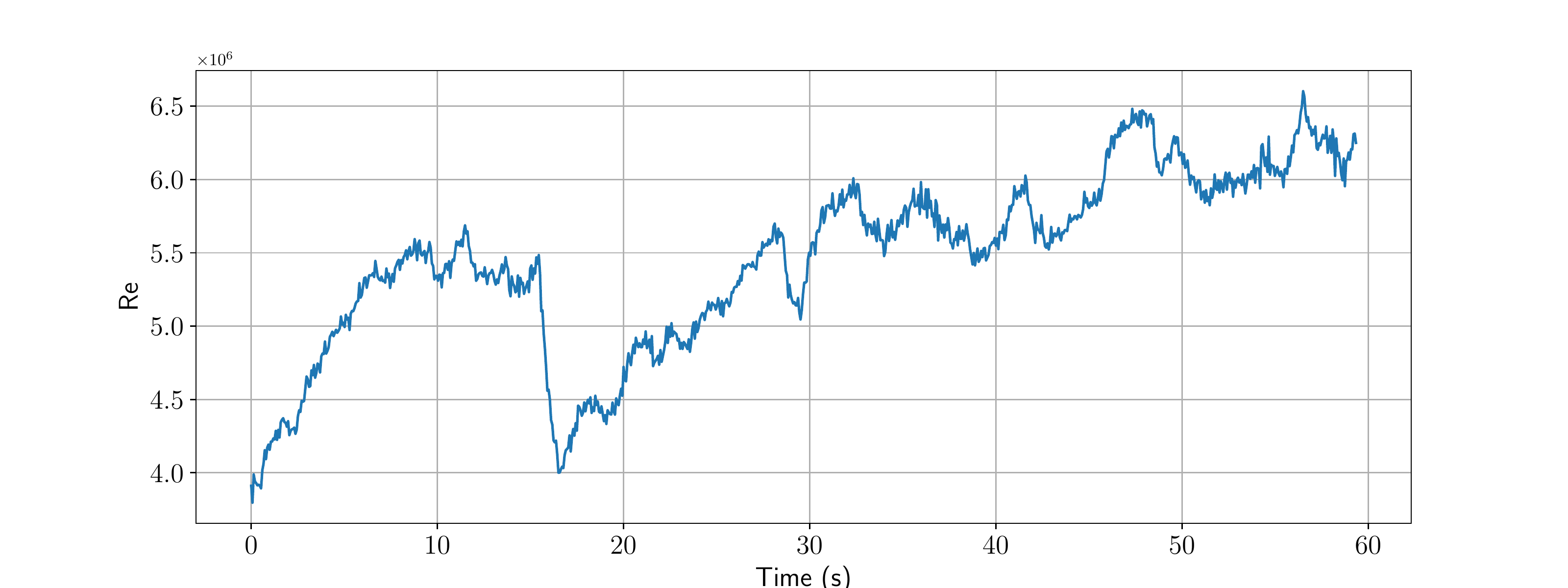}
    \caption{Reynolds number ($Re$) evolution. In all of our wind simulations, the value of $Re$ increases as the wind becomes more turbulent.}
    \label{fig:Re_evolution}
\end{figure}

\customsubsection{Experimental Setup:}\label{subsec:experimental_setup}
\note{Environment}
We conduct all our experiments in a 2-dimensional square simulation space $\workspace$ of size $10\times 10$ sq m. We simulate $M=60$ wind fields $\wind$ by solving the Navier-Stokes equations inside $\workspace$, with random initial conditions. Each $\wind$ is guaranteed to be in a turbulent regime at $Re \geq \expnumber{4}{3}$. The turbulence intensifies with time in all of our $\wind$, increasing the $Re$ value as shown in Fig. \ref{fig:Re_evolution}. We control the maximum possible wind speed in each wind simulation and bound it to a value of $15$ m/s. We generate the wind flows using a publicly available Computational Fluid Dynamics (CFD) software~\cite{fluiddyn,fluidfft,fluidsim}. We provide a script to compute simulations along with the project's source code\footnote{\url{https://github.com/dipaco/robot_wind_navigation}}.
\note{Describe the robots}
For each robot, we compute the drag force exerted by the wind as per \eqref{eq:drag}. We set the air density to $\rho=1.184$ \si{kg . m^{-3}} and the drag coefficient to $C_d=0.47$. Additionally, we assume all of the robots are small spheres of radius $r=0.1$ \si{m} with a cross-sectional area of $A=\pi r^2$ sq m. We use lattice formations in all of our experiments at different sizes and chose the lattices' initial location to fit entirely into $\workspace$.

\note{Training recipe}
We train all our models on only $50$ of the wind simulations and reserve the remaining $10$ for testing. We train each RL model for $\expnumber{5}{6}$ steps using a replay buffer of $\expnumber{2}{5}$. This replay buffer's size ensures the RL model focuses more on recent experiences where the reward is expected to be better. We optimize the SAC's loss functions from \eqref{eq:J_Q} and \eqref{eq:J_pi} using Adam optimizer~\cite{Kingma2015AdamAM} with a fix learning rate of $\expnumber{1}{-3}$. At training, all the episodes have a fixed duration of $T=60$ s. We set the weights in the reward to $\boldsymbol{\beta} = [1, 1, 10, 10].$ We use the $k$-nearest neighbor algorithm (knn) to define the graph's adjacency matrix at each time step. In all of our experiments, we start the robot's formation at random locations within $\workspace$. We report average absolute errors over $20$ episodes with corresponding $95\%$ accuracy confidence intervals.

\begin{figure}
    \setlength\belowcaptionskip{-1.0\baselineskip}
     \centering
     \begin{subfigure}[b]{0.47\textwidth}
        \setlength\belowcaptionskip{0.0\baselineskip}
         \centering
         \includegraphics[width=\textwidth, trim=0.0cm 0.0cm 0.0cm 0.0cm,clip]{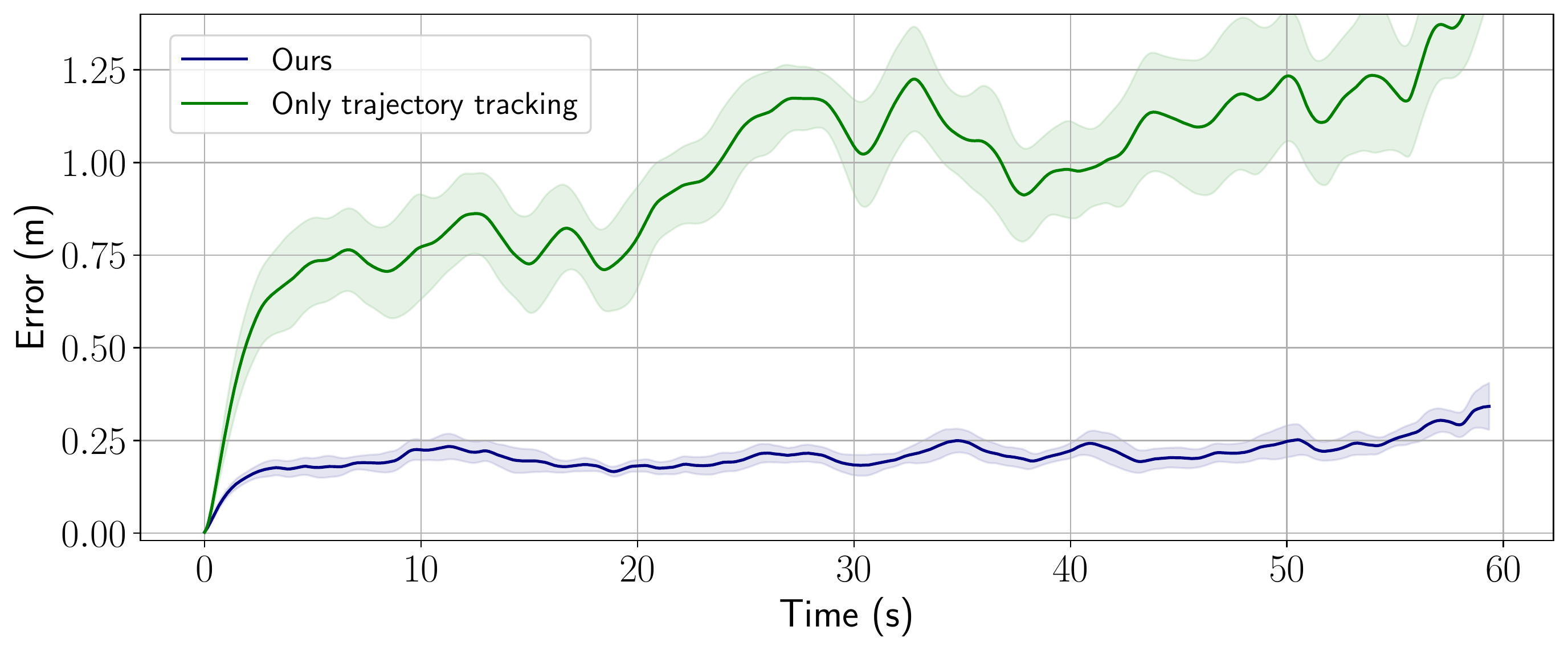}
         \caption{Position error}
         \label{fig:y equals x}
     \end{subfigure}
     \hfill
     \begin{subfigure}[b]{0.47\textwidth}
        \setlength\belowcaptionskip{0.0\baselineskip}
         \centering
         \includegraphics[width=\textwidth, trim=0.0cm 0.0cm 0.0cm 0.0cm,clip]{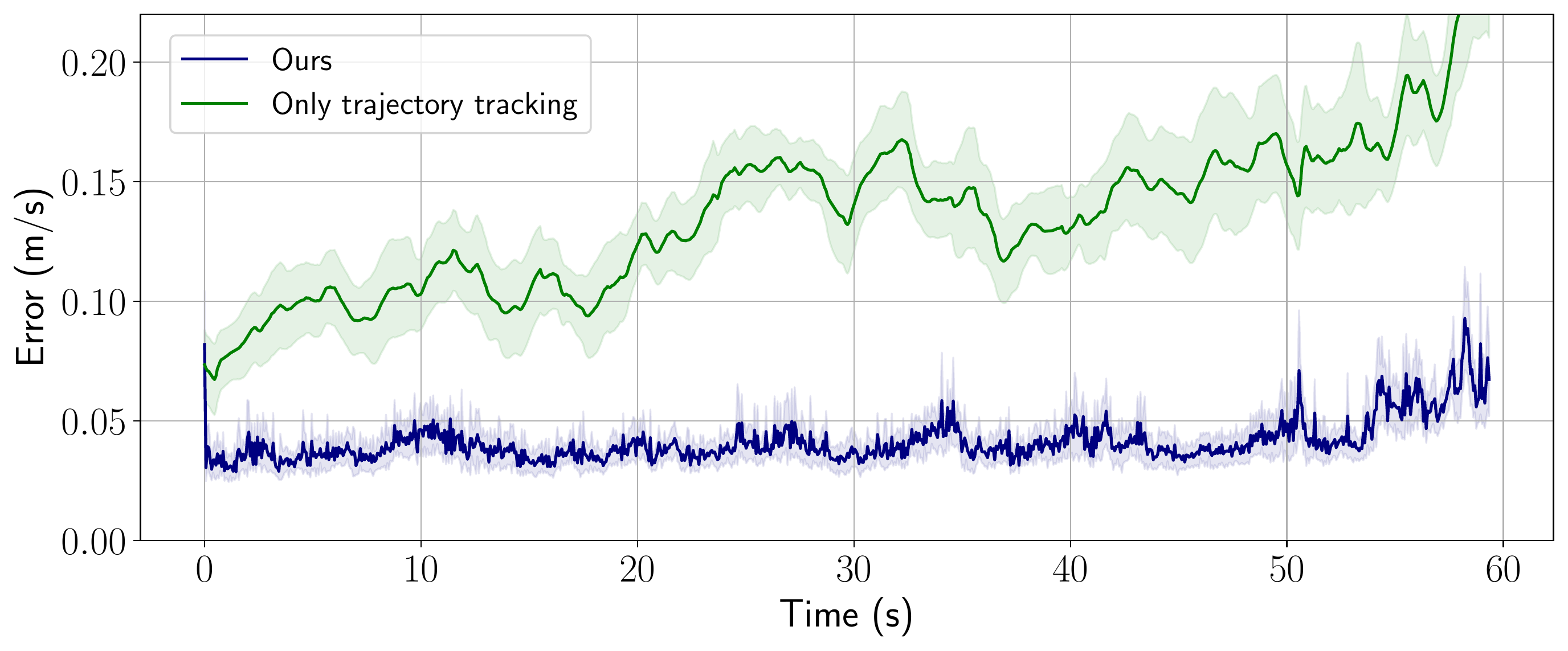}
         \caption{Velocity error}
         \label{fig:three sin x}
     \end{subfigure}
        \caption{Our method's performance compared to only the trajectory-tracking control. The curves show the mean error across $20$ episodes with corresponding $95\%$ confidence interval.}
        \label{fig:pd_vs_rl}
\end{figure}

\customsubsection{Experiment 1: Wind compensation.}
In this experiment, we explore the benefits of assisting the trajectory-tracking control from \eqref{eq:u_tt} with our RL method to compensate for the force that a turbulent wind field exerts on a robot. To this end, we compute the position and velocity errors at each time $\tau$ of the trajectory-tracking control with and without the RL wind-compensation strategy. We use a formation size of $n=25$ robots and a neighborhood size of $k=12$. We report average errors over $20$ episodes and all $n$ robots in the swarm and summarize the results in Fig.~\ref{fig:pd_vs_rl}. The noise to all our sensors follows a zero-mean Gaussian distribution with $\sigma = 0.001$ for the position and velocity sensor and $\sigma = 0.1$ for the pressure sensor.

\begin{figure}
    \setlength\belowcaptionskip{-0.5\baselineskip}
    \centering
    \includegraphics[width=\linewidth, trim=0.0cm 0.0cm 0.0cm 0.0cm,clip]{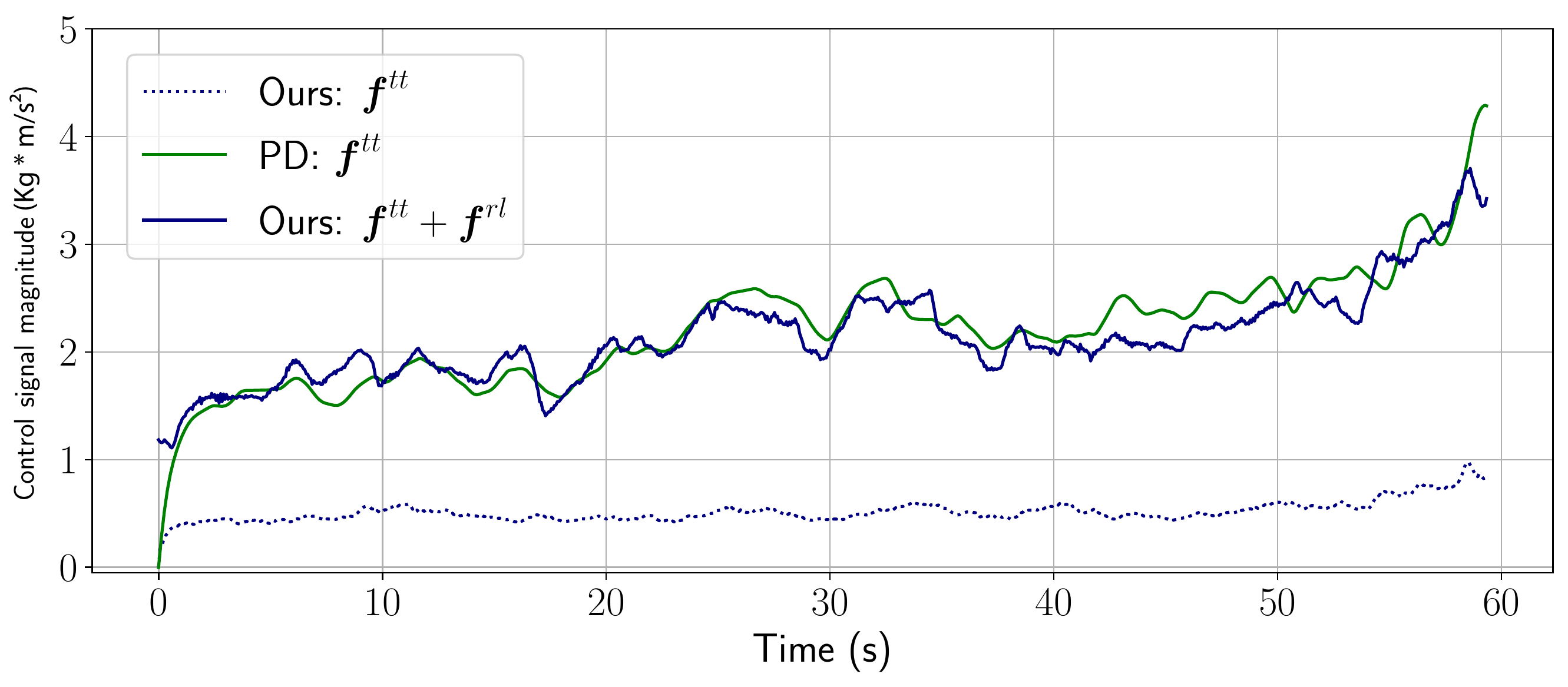}
    \caption{Magnitude of the control signal. Solid lines show the total action signal for our method (blue) and only the trajectory-tracking control (green). Additionally, we show the isolated tracking-trajectory component of our method (dotted blue).}
    \label{fig:control_signal_pd_vs_rl}
\end{figure}

Our method (blue curve) shows a statistically significant improvement compared to trajectory tracking only (green curve). Note that our method maintains the position and velocity errors at relatively stable values despite the increase in the turbulence regime described in Fig. \ref{fig:Re_evolution}. From this result, we conclude that our proposed method can capture and compensate for the wind effects that affect the robots, regardless of the intensity and complexity of the wind.
Additionally, we report in Fig.~\ref{fig:control_signal_pd_vs_rl} the magnitude of the total control signal of each robot -- in Newtons -- averaged over all the robots in the formation. Recall that the total control signal from our proposed method is the sum of trajectory-tracking control and the RL action as per~\eqref{eq:forces}. The magnitude of the control signal is associated with the amount of energy the robots use to complete their task, e.g., tracking a trajectory. By comparing the curves in Fig. \ref{fig:control_signal_pd_vs_rl}, we conclude that our method achieves significantly lower errors with approximately the same control signal magnitude. Hence, our methods preserve the amount of energy the robots use to fulfill their tasks while achieving better performance. Moreover, we report the trajectory-tracking component of our method (dotted blue) and highlight the smoothness of the curve compared to the trajectory-tracking alone. We conclude that this occurs because the robots can track a target free of perturbations when the RL compensates for the wind's effect. 

\customsubsection{Experiment 2: Sensitivity Analysis.}
In this experiment, we test our method's sensitivity regarding two key parameters of our model: the robot's neighborhood size, $k$, and the number of robots in the team, $n$.

\begin{figure}
    \setlength\belowcaptionskip{-0.7\baselineskip}
    \centering
    \includegraphics[width=\linewidth, trim=0.0cm 0.0cm 0.0cm 0.0cm,clip]{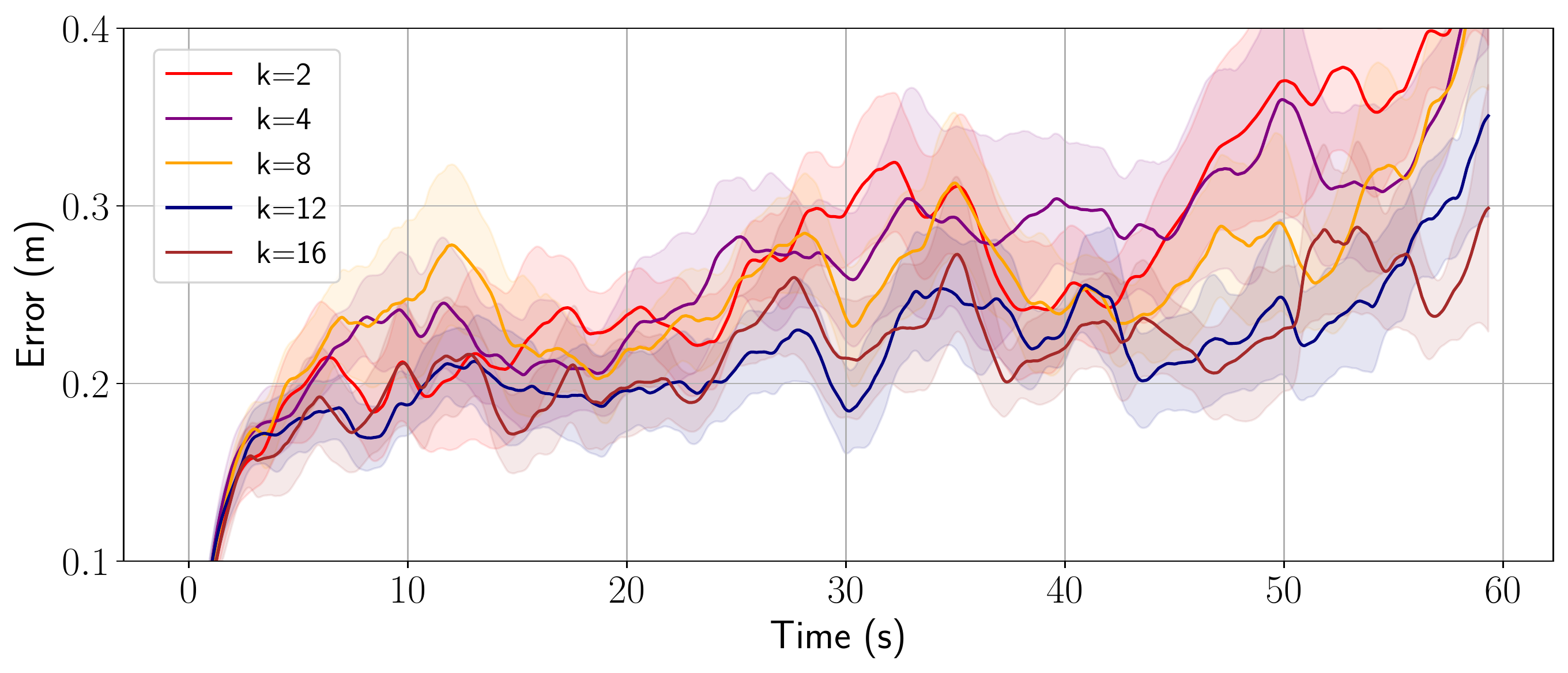}
    \caption{Sensitivity to the neighborhood size.}
    \label{fig:sensitivity_to_k}
\end{figure}

We investigate the effect of the neighbor size on our method's ability to learn a wind compensation action. To this end, we train five models varying the neighborhood size at increasing values of $k$, such that $k \in \{2, 4, 8, 12, 16\}$, and maintain the formation size constant at $n=25$. We report the average position error of each of these models in Fig. \ref{fig:sensitivity_to_k}. Our results show a decrease in the error when  $k$ increases. Note that the error gap between curves with lower values of $k$ and curves with larger $k$ increases with the turbulence intensity (See Fig. \ref{fig:Re_evolution}). We did not observe a significant improvement in performance for models trained with $k > 12$.

\begin{figure}
    \setlength\belowcaptionskip{-0.5\baselineskip}
    \centering
    \includegraphics[width=0.85\linewidth, trim=0.0cm 0.0cm 0.0cm 0cm,clip]{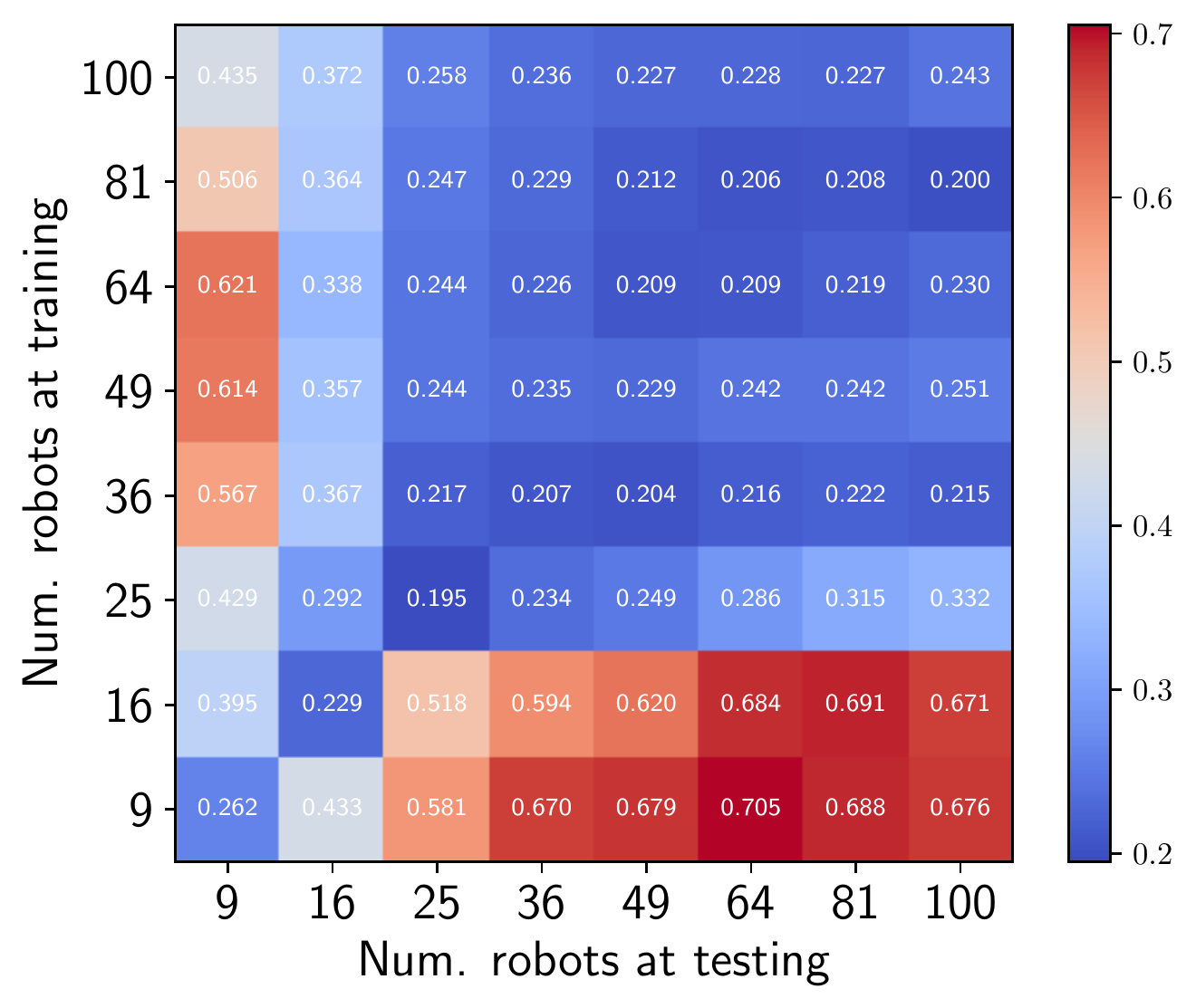}
    \caption{Sensitivity to the formation size.}
    \label{fig:sensitivity_to_formation_size}
\end{figure}

We train eight of our RL-based models, varying the training and testing formation size to test our method's sensitivity to the training formation. We use $n^{\textrm{train}}, n^{\textrm{test}} \in \{3^{2}, ..., 10^{2}\}$ while maintaining the neighborhood size constant at $k=12$. We report the average position error of each test in Fig. \ref{fig:sensitivity_to_formation_size}. Note that our method scales well to large formation when trained with enough robots without retraining, e.g., $n \geq 25$. The performance decrease in the first two columns results from testing on formations that do not satisfy the neighborhood requirements when training the models, $k=12$. Similarly, the two first rows in Fig. \ref{fig:sensitivity_to_formation_size} show a decrease in performance due to training with insufficient robots. In this last scenario,  the neighborhood cannot meet the requirements to capture the wind dynamics.

\customsubsection{Experiment 3: Ablation Study.}
We conduct an ablation study to investigate the contribution of our proposed architecture to the overall system. We compare our model with five baselines to highlight the advantages of information sharing in our model.
%
In the first baseline, we replaced the GCNN with an MLP shared across all robots in the team. The MLP has the same number of hidden layers and neurons but does not share information with its neighbors. It can only access the features of the nodes in which it is operating. The second baseline is a deeper MLP of four hidden layers. The increase in depth has the effect of approximately doubling the number of weights. Similarly, the third baseline is a wider MLP with a layer width of $128$ neurons. Doubling the layer's width increases the number of weights in the base MLP by approximately a factor of four.

\setlength{\tabcolsep}{3.5pt} 
\renewcommand{\arraystretch}{1.1} 
\begin{table*}[t]
    \setlength\belowcaptionskip{-1.2\baselineskip}
    \centering
    \begin{tabular}{p{80pt}cccccccc}
     \textbf{Method} & \multicolumn{5}{c}{\textbf{Position Error}} \\
     \toprule
     & \textbf{Time:} & $0s$ & $10s$ &  $20s$  & $30s$  & $40$  & $50s$ & $60s$ \\
     & $\mathbf{Re}$: & $\expnumber{3.9}{6}$ & $\expnumber{4.3}{6}$ &  $\expnumber{4.4}{6}$  & $\expnumber{5.6}{6}$  & $\expnumber{5.2}{6}$  & $\expnumber{5.3}{6}$ & $\expnumber{6.6}{6}$ \\
     \toprule
     Base MLP &   & 0.092 $\pm$ 0.022 & 0.362 $\pm$ 0.065 & 0.338 $\pm$ 0.048 & 0.419 $\pm$ 0.072 & 0.414 $\pm$ 0.050 & 0.418 $\pm$ 0.073 & 0.591 $\pm$ 0.085 \\
     Wider MLP &   & 0.105 $\pm$ 0.022 & 0.418 $\pm$ 0.076 & 0.340 $\pm$ 0.046 & 0.533 $\pm$ 0.053 & 0.405 $\pm$ 0.047 & 0.431 $\pm$ 0.051 & 0.607 $\pm$ 0.079 \\
     Deeper MLP &   & \textbf{0.089 $\pm$ 0.014} & 0.367 $\pm$ 0.068 & 0.282 $\pm$ 0.051 & 0.438 $\pm$ 0.049 & 0.410 $\pm$ 0.039 & 0.457 $\pm$ 0.071 & 0.631 $\pm$ 0.084 \\
     Only trajectory tracking &   & 0.282 $\pm$ 0.055 & 0.803 $\pm$ 0.156 & 0.699 $\pm$ 0.063 & 1.020 $\pm$ 0.132 & 0.939 $\pm$ 0.131 & 0.932 $\pm$ 0.152 & 1.325 $\pm$ 0.184 \\
     \bottomrule
     Ours - No rel. position &   & 0.092 $\pm$ 0.011 & 0.287 $\pm$ 0.058 & 0.235 $\pm$ 0.030 & 0.325 $\pm$ 0.052 & 0.316 $\pm$ 0.043 & 0.340 $\pm$ 0.059 & 0.509 $\pm$ 0.081 \\
     Ours - Full model &   & 0.128 $\pm$ 0.021 & \textbf{0.194 $\pm$ 0.027} & \textbf{0.173 $\pm$ 0.014} & \textbf{0.191 $\pm$ 0.025} & \textbf{0.228 $\pm$ 0.017} & \textbf{0.241 $\pm$ 0.041} & \textbf{0.311 $\pm$ 0.059} \\
    
    \end{tabular}
    \caption{Quantitative results of the ablation study. The table shows the average position error along the duration of an episode. We report the average $Re$ at selected times.}
    \label{tab:ablation}
\end{table*}

We include a fourth baseline to test the ability of our model to learn spatially distributed information from a robot's neighbor. In this baseline, we ablate the inclusion of the relative position $\boldsymbol{r}_{i,j}$ in the convolution definition of \eqref{eq:graph_conv}. By removing the relative position, our GCNN can still share information between a robot $i$ and its neighbors. However, the robot cannot identify where those neighbors are located relative to itself. Finally, the last baseline is the trajectory-tracking controller without our RL wind compensation.
\begin{figure}
    \setlength\belowcaptionskip{-1.5\baselineskip}
    \centering
    \includegraphics[width=\linewidth, trim=0.0cm 0.0cm 0.0cm 0.cm,clip]{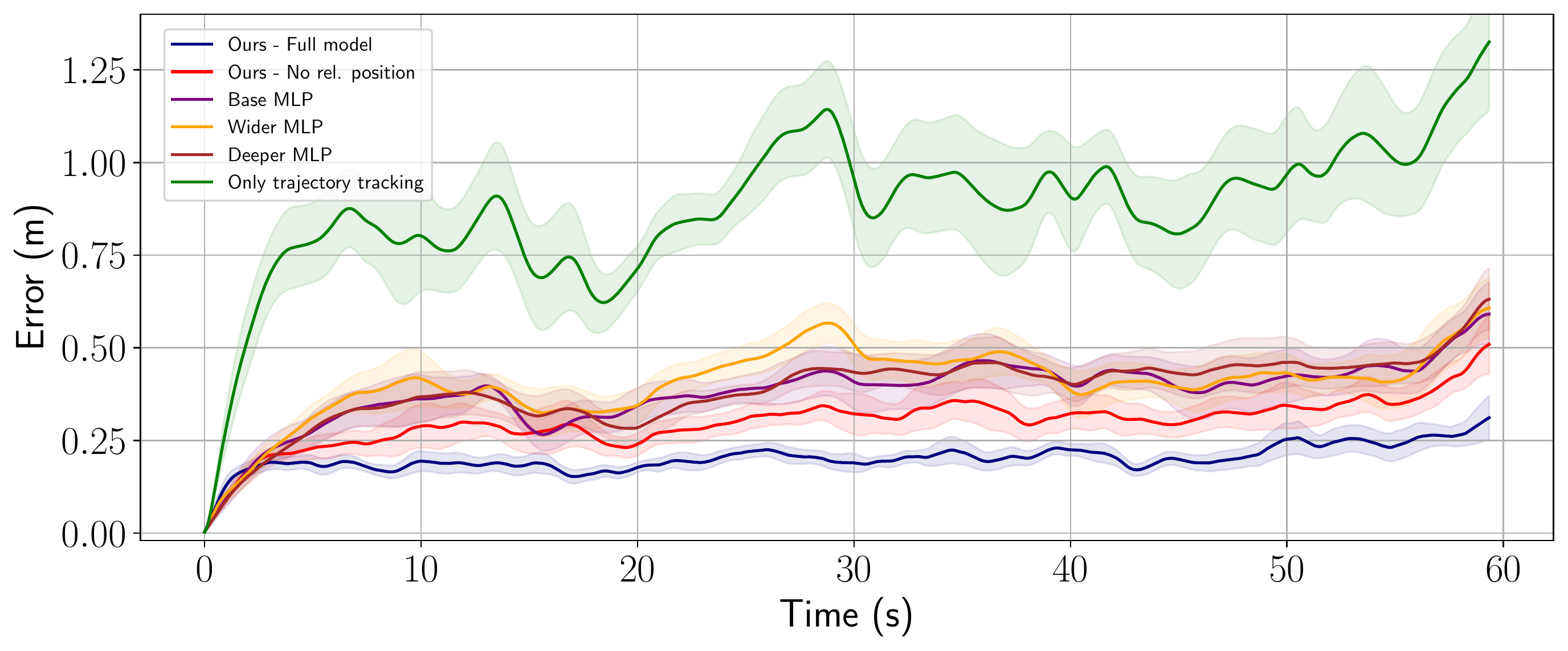}
    \caption{Ablation study results. Our method (blue curve) shows a statistically significant improvement compared to all baselines.}
    \label{fig:ablation}
\end{figure}

Our experiments show that our approach achieves the lowest position error among all methods in the ablation study. We summarize all the ablation experiments in Tab. \ref{tab:ablation} and Fig.~\ref{fig:ablation}. We report the average position error of each method along the corresponding $Re$ values along an episode. Note that all the MLP-based baselines have similar error curves, despite the significant increase in capacity of the Deeper and Wider MLP. These results demonstrate that the advantages of our method arise from our GCNN-based RL strategy and not from the neural network's size.

\customsubsection{Discussion:}
Navigation in turbulent flows with high levels of turbulence, $Re > \expnumber{4}{6}$, is a challenging problem. However, these high turbulence levels have not been studied in the state-of-the-art. This scenario is especially challenging for a single robot since its perception of the flow is limited. In this paper, we leveraged multiple robots to navigate high-turbulence flows and evaluate different factors that help understand the difficulties of operation in this type of aggressive environment. Although the spherical robots we presented only exist in simulations, our method can be implemented in actual robots.

\section{Conclusion and Future Work}
In this paper, we introduced a novel RL-based method to control a team of aerial robots to track a trajectory while working together in a dynamic, turbulent wind field. Our method's strategy decouples the trajectory-tracking controller and wind compensation. So our method can learn to compensate for the wind turbulence independently of the motion controller.
Our RL approach allowed us to find an optimal policy to compensate for the wind force via a graph neural network designed to share information among the robotic team members. Our method shows that sharing sensor measurements between nearby robots provides valuable information to improve the robots' turbulence compensation and learn spatially-distributed wind patterns. We demonstrate the advantages of our strategy through several simulations strategically designed to test our method's performance for wind compensation, its scalability to large robot formations, and its parameter sensitivity. 

In future work, we want to design and implement a lab testbed to generate air flows with high turbulence levels like the ones presented in this paper. Although this type of testbed has a high cost and complexity, it would allow us to test and extend methods for navigation in high turbulence. Another direction of future work is to test our model against increasing sensor noise -- as it can arise from more turbulent winds. Additionally, we want to model the temporal dependencies of turbulent vector fields through recurrent neural network architectures such as GRU or LSTM.

\ifCLASSOPTIONcaptionsoff
  \newpage
\fi



%
\bibliographystyle{IEEEtran}
\bibliography{camera_ready}

%




\end{document}